\documentclass[runningheads]{llncs}

\usepackage{eccv}

\usepackage{geometry}
\usepackage{eccvabbrv}
\usepackage{microtype}
\usepackage{graphicx}
\usepackage{booktabs} % for professional tables
\usepackage{wrapfig}
\usepackage{hyperref}

\usepackage{amsmath}
\usepackage{amssymb}
\usepackage{mathtools}
\usepackage{hyperref}
\hypersetup{
    colorlinks=true,
    linkcolor=blue,
    filecolor=magenta,      
    urlcolor=cyan,
    pdftitle={Overleaf Example},
    pdfpagemode=FullScreen,
    }
\usepackage{caption}
\usepackage{xcolor}
\usepackage{derivative}
\usepackage[dvipsnames]{xcolor}
\usepackage[capitalize,noabbrev]{cleveref}
\usepackage{graphicx}
\usepackage{booktabs}
\usepackage{multirow}
\usepackage[accsupp]{axessibility}  % Improves PDF readability for those with disabilities.
\usepackage{hyperref}

\usepackage{orcidlink}
\usepackage{float}   % for [H]
\usepackage[bottom]{footmisc}
\begin{document}

\title{FEEL (Force-Enhanced Egocentric Learning): \\ A Dataset for Physical Action Understanding}
\titlerunning{FEEL}

\author{
Eadom Dessalene \and
Botao He \and
Michael Maynord \and
Yonatan Tussa \and
Pavan Mantripragada \and
Yianni Karabatis \and
Nirupam Roy \and
Yiannis Aloimonos
}

\authorrunning{Dessalene et al.}

\institute{
University of Maryland, College Park, USA\\
\email{Email: edessale@umd.edu}
}


\maketitle
\vspace{-0.4em}
\begin{center}
\small
\href{https://www.cs.umd.edu/~edessale/feel}{\texttt{https://www.cs.umd.edu/\textasciitilde edessale/feel}}
\end{center}
\vspace{-0.8cm} % try -0.6cm to -1.8cm
\begin{figure}[h]
    \centering
    \includegraphics[width=0.9\textwidth]{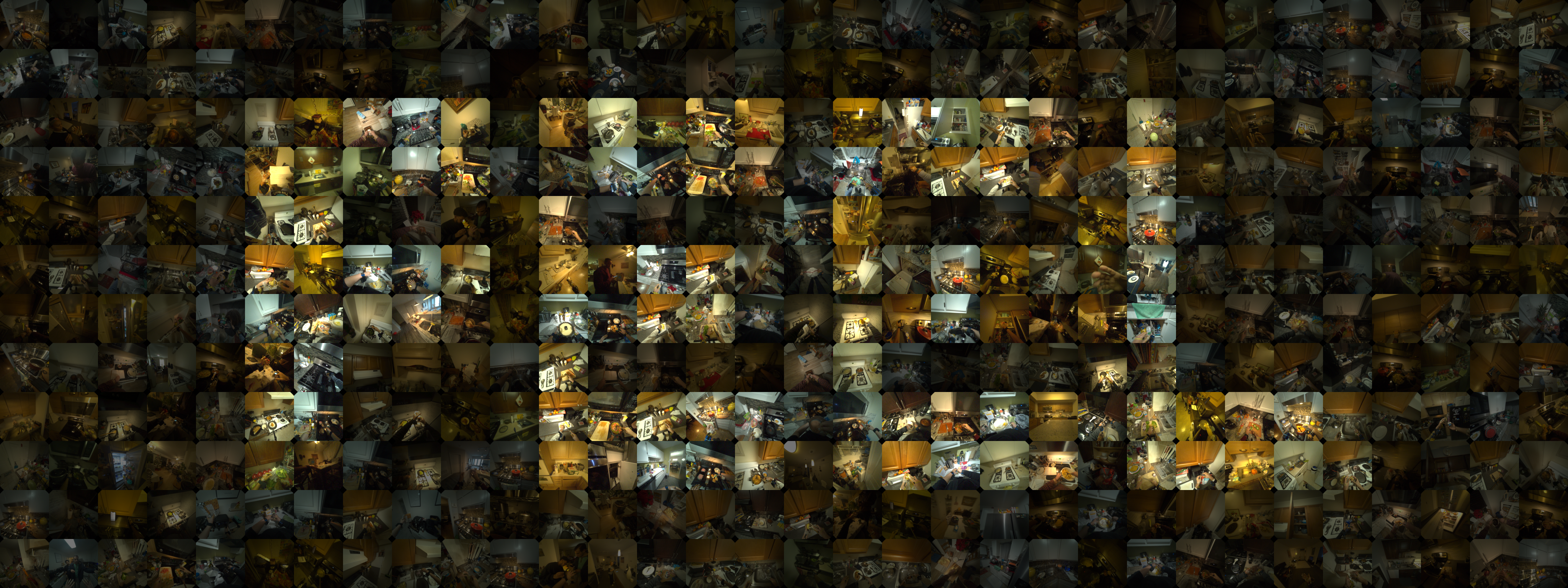}
   \caption{\textbf{FEEL} pairs egocentric video with force measurements to capture the physical causes—not just visual effects—of hand-object interaction.}
    \label{fig:data_mosaic}
\end{figure}

% \footnote{Example video and force measurements from the dataset can be found \href{https://drive.google.com/drive/folders/1jowAwaqd71aAZDpDnatxEQ_yfWCg4cAo?usp=sharing}{here}}

\begin{abstract}
  We introduce FEEL (Force-Enhanced Egocentric Learning), the first large-scale dataset pairing force measurements gathered from custom piezoresistive gloves with egocentric video. Our gloves enable scalable data collection, and FEEL contains approximately 3 million force-synchronized frames of natural unscripted manipulation in kitchen environments, with $\sim$45\% of frames involving hand-object contact. Because force is the underlying cause that drives physical interaction, it is a critical primitive for physical action understanding. We demonstrate the utility of force for physical action understanding through application of FEEL to two families of tasks:  (1) \textbf{contact understanding}, where we jointly perform temporal contact segmentation and pixel-level contacted object segmentation; and, (2) \textbf{action representation learning}, where force prediction serves as a self-supervised pretraining objective for video backbones. We achieve state-of-the-art temporal contact segmentation results and competitive pixel-level segmentation results without \textit{any} need for manual contacted object segmentation annotations. Furthermore we demonstrate that action representation learning with FEEL improves transfer performance on action understanding tasks without \textit{any manual labels} over EPIC-Kitchens, SomethingSomething-V2, EgoExo4D and Meccano.
\end{abstract}

\section{Introduction}
\label{sec:introduction}

\begin{figure}[t!]
    \centering
    \includegraphics[width=1\textwidth]{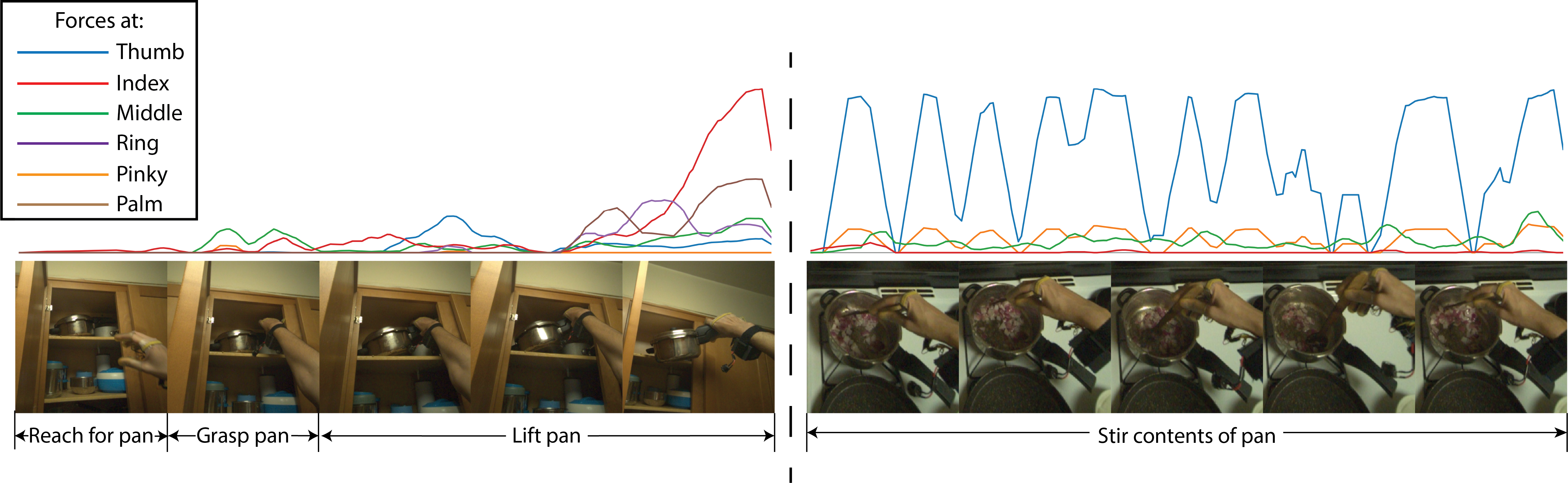}
    \caption{\textbf{Forces reveal interaction dynamics.} Representative sequences showing how force measurements disambiguate physical interactions. \textbf{Left:} Grasping and lifting a pan exhibits distinct force signatures with an onset at the end of the reach and a spike as the pan is lifted.  \textbf{Right:} Stirring produces rhythmic force patterns reflecting cyclic manipulation.}
    %fewer images, larger images, fix cropping, grasp pan and lift pan for first sequence, remove white line going through second seg of images, make images more tightly cropped, convert take teapot into grasp, find cleaner example than second row because there is noise during release, legend on top left, prevent overlap of action colors on force plots
    %put two figures side by side, remove middle sequence, make labels align with figure boundaries, remove color coding for action boundaries, add colon to sensor locations and box around entire legend (including sensor locations)
    %dotted line not centered, label forces by changing legend to "Forces at:"
    \label{fig:data_sequence}
\end{figure}

Physical manipulation actions are defined by the forces the hands exert on the world. For this reason, force should not be treated as a peripheral signal, but as a primitive for physical action understanding. Vision, in contrast, primarily measures motion and appearance — observable effects of action but not causal in themselves. Learning actions from video alone leaves the underlying action structure implicit in a high dimensional stream of pixels - incorporating force turns the hidden action structure into an intrinsically lower dimensional learnable signal.

\begin{figure}[t!]
    \centering
    \includegraphics[width=1\textwidth]{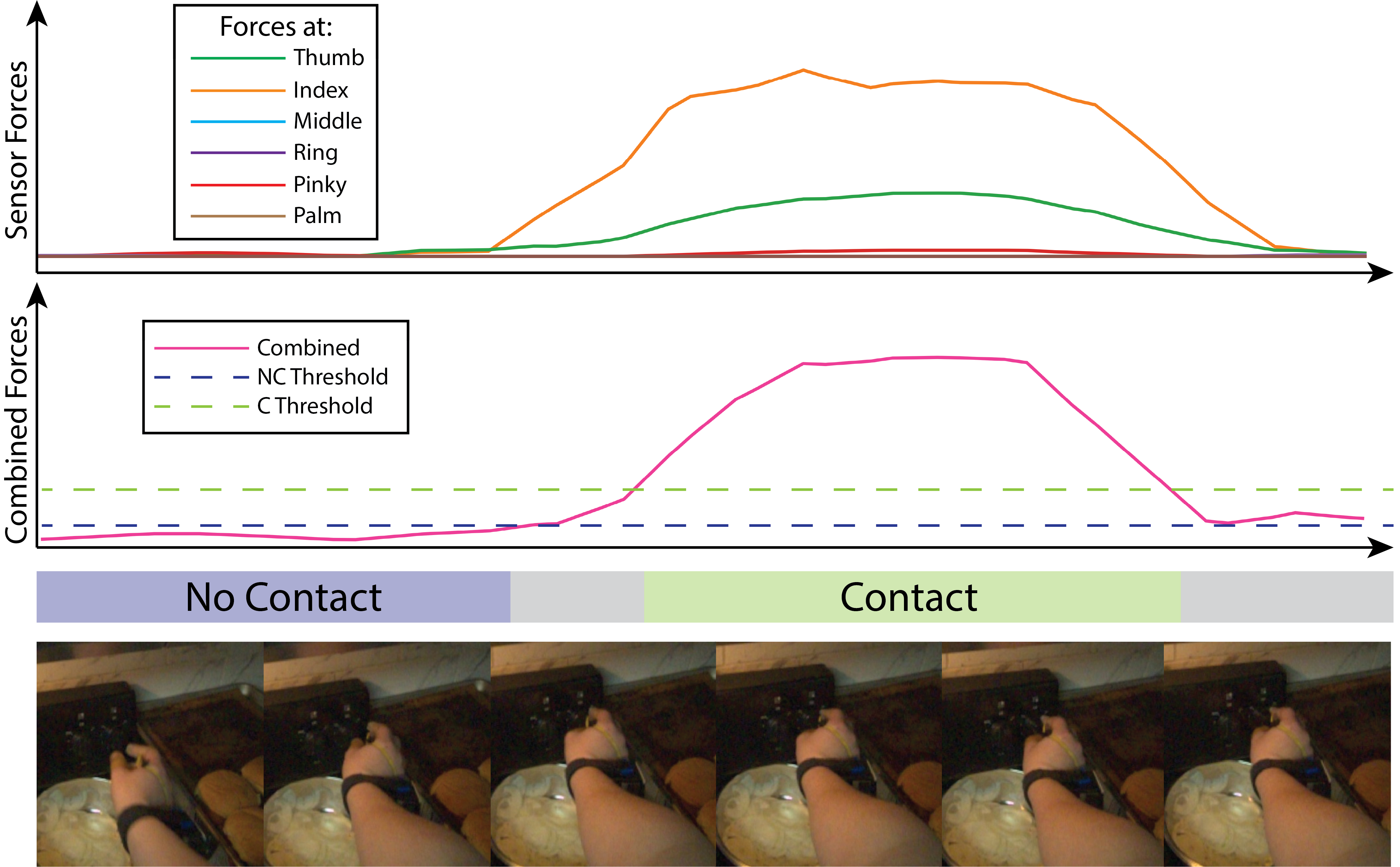}
    % \caption{Data sequence, 3 row. Two column figure. 4 columns in the figure, different time points in the same short clip. Make it two column figure - pick consecutive images that are visually similar but different in force readings}
    \caption{\textbf{Contact detection from force measurements.} Despite visual similarity across frames, force measurements precisely identify contact boundaries. \textbf{Top:} Raw sensor forces (pinky and middle finger dominate this grasp). \textbf{Middle:} Consolidated force with dual thresholds—above C threshold indicates contact, below NC threshold indicates non-contact, between is ambiguous (excluded). Forces enable scalable contact supervision without manual annotation of ambiguous visual transitions.}
    %4 images not 5, crop hand and knob, change color of third plot and add dotted line for contact threshold and vertical dotted line to segment rgb frames (and show span of contact/no contact), labels on x and y axis and ticks and arrows on x and y axis
    %add 4 more images, make images bigger, make threhsold line dotted, bring down threshold line, make frames on bottom, with nc/c above, and nc/c frame aligned
    %Remove grey line, incorporate two more sensors into top plot, make nc and c threshold narrower, sensor locations -> sensor forces, make change from figure 2 legend to keep legend identical, vertical axes, forces -> sensor forces, consolidated forces, expand the line plots to span all of time axes
    %Consolidated forces -> consolidated force (axes and legend), remove header for legend, overlap of colors between first and second plots, no contact and contact boundaries do not align with points of intersection, separate annotations from plots+rgb, reduce fontsize of contact no contact, shift legend to the left instead of to the right, make spacing within the legends equally spaced, consolidated -> combined (y axes and legend)
    %add space between contact annotations and rgb frames/plots, make the box larger vertically, reverse odering of nc threshold and c threshold, combined forces -> combined force, remove title from legend of first plot, left justification s otop and bottom legend align
    \label{fig:ambiguous}
\end{figure}

Egocentric video datasets have enabled major progress by focusing on the visual modality for action understanding tasks. This data is useful, but it lacks a modality directly capturing the causal dynamics of action: force. Force is absent from existing datasets largely because it cannot be reliably obtained through manual labeling. The physical response of objects to forces applied are often very subtle and difficult for annotators to recognize directly from visual changes. This difficulty includes annotating the making and breaking of contact. For illustration, see Figure \ref{fig:ambiguous} for a case where appearance is inconclusive while force disambiguates.

In order to address the shortcomings of existing datasets, we introduce FEEL (Force-Enhanced Egocentric Learning). FEEL is built around a custom force sensing glove (see Figure \ref{fig:hardware}) consisting of $5$ Piezoresistive sensors mounted to the pads of the fingers, as well as a long Piezoresistive sensor mounted to the palm. This glove when worn provides a stream of measurements of the forces imparted by hands. We record this stream alongside egocentric captures collected with Meta Aria glasses \cite{engel2023project}. The result is time-synchronized egocentric video, audio, and force sensing of natural, unscripted, unstructured manipulation. See Figure \ref{fig:data_sequence} for illustration of synchronized video and force sensing streams. Contact is derived from force exchange, yielding temporally precise per-frame contact state and boundaries (see Figure \ref{fig:ambiguous}). A key benefit to force-derived contact labels as opposed to manual annotation of contact labels over RGB frames is scalability. We focus on kitchens, which provide a broad range of interaction/action types. FEEL contains $\sim 3$M frames over $\sim 27$ hours of dense manipulation, with $\sim 45\%$ of frames involving contact.

% Our dataset is built around a custom force sensing glove: we create a glove consisting of $5$ Piezoresistive sensors mounted to the pads of the fingers, as well as a long Piezoresistive sensor mounted to the palm. This glove when worn provides a stream of measurements of the forces imparted by hands. We record this stream alongside egocentric captures collected with Meta Aria glasses; the egocentric view aligns sensing with the physical origin of force (the hands), yielding time-synchronized egocentric video, audio, and force sensing during natural, unscripted, unstructured manipulation. Contact is derived from force exchange, yielding temporally precise per-frame contact state and boundaries (See Figure {\color{red} 2} for illustration) that would otherwise be difficult to annotate reliably from RGB alone. A key benefit to contact labels derived from forces as opposed to manual annotation is scalability {\color{blue} expand}. We focus on kitchens, which provide a broad range of interaction/action types for hand manipulation modeling. Overall, we collect $\sim 2$M frames over $\sim 17$ hours of dense manipulation, with $\sim 45\%$ of frames involving contact. 

Using force as supervision provides a low-dimensional physically grounded signal that constrains how action is extracted from high-dimensional pixel space. Along these lines we develop learning methods leveraging forces as supervision for the following tasks: Contact Understanding and Action Representation Learning (see Figure \ref{fig:task_setup}). 

% Force supervision provides a low dimensional, physically grounded signal that helps extract action from high dimensional pixel observations. To leverage the force supervision, we developed two methods for two tasks, Contact Understanding and Action Representation Learning (see Figure \ref{fig:task_setup}), and achieve state of the art performance on both.

% {\color{blue} (Give tasks names like in the intro + experiments)}. 

In the \textbf{Contact Understanding} task we leverage our force signal in the modeling of contact in egocentric video. This task is defined as jointly: (1) temporal contact segmentation - classifying the per-frame contact state over time for each hand; and, (2) pixel-level segmentation - segmenting the object(s) in contact with each hand. Our force measurements provide temporally precise supervision for the onset and offset of contact, enabling the learning of sharp interaction boundaries which would be difficult to learn over imprecise contact annotations (see the Supplementary Materials for examples).

% Using force as supervision provides a low-dimensional structural signal that constrains how action structure is extracted from high-dimensional pixel space. Along these lines we introduce two learning tasks leveraging forces as supervision. In the \textbf{first} task we leverage our force signal in the modeling of contact in egocentric video. This task is defined as jointly: (1) temporal contact segmentation - classifying the per-frame contact state over time for each hand; and, (2) pixel-level segmentation - segmenting the object(s) in contact with each hand. Our force measurements provide temporally precise supervision for the onset and offset of contact, enabling the learning of sharp interaction boundaries which would be difficult to learn over imprecise contact annotations [].

In the \textbf{Action Representation Learning} task, we use FEEL to demonstrate the utility of force for learning action representations over video. Training to predict forces encourages the network to infer physically grounded action structure from video. We evaluate transfer of the resulting representations to the task of action recognition. We evaluate on established benchmarks, including EPIC-Kitchens, Ego-Exo4D, Something-Something V2, and Meccano.

The primary contributions of this work are as follows:
\begin{enumerate}
    \item \textbf{Release of FEEL: } We release the largest egocentric force–video dataset, capturing the physical causes - not just the visual effects - of hand-object manipulation. FEEL consists of approximately $3$ million frames collected during natural, unscripted manipulation.
    \item \textbf{Scalable force-derived contact labels: } We derive per-frame contact states, releasing $1.35M$ frames of hands and objects in contact (along with object segmentations) and $1.65M$ frames of hands and objects not in contact.
    \item \textbf{Self-supervised force pre-training pipeline} A demonstration of the benefits of training force-aware video backbones for downstream tasks. Force-aware pretraining over FEEL improves downstream action recognition across nearly all datasets.
    \item \textbf{A new state-of-the-art contact estimation model: } We train a model to jointly perform temporal contact segmentation and pixel-level object-in-contact segmentation over still images. Our model achieves state-of-the-art temporal contact detection across all three benchmarks.

\end{enumerate}

The rest of this paper is structured as follows: In Section \ref{sec:related_work} we cover related work, in Section \ref{sec:methods} we detail our method, in Section \ref{sec:experiments} we cover evaluations, and in Section \ref{sec:discussion} we discuss and conclude.

\section{Related Work}
\label{sec:related_work}
% Blah
\textbf{Contact Understanding}
The predominant paradigm for contact understanding relies on large, manually annotated datasets over images \cite{shan2020understanding,narasimhaswamy2020detecting,cheng2023towards,chen2023detecting,tripathi2023deco} or video \cite{darkhalil2022epic,cheng2023towards,dessalene2021forecasting,liu2022joint}, which incur significant annotation cost and label noise. Some works mitigate this through semi-automatic segmentation workflows that propagate manual annotations using off-the-shelf video tools \cite{darkhalil2022epic,dessalene2021forecasting}. Alternative approaches collect data in controlled lab settings with motion capture to precisely track hands and objects in 3D \cite{brahmbhatt2020contactpose,fan2023arctic,taheri2020grab}, or sidestep real-world collection entirely via synthetic data \cite{leonardi2024synthetic}.

\textbf{Force Understanding}
Imbuing vision models with physical understanding remains highly challenging and underexplored \cite{fermuller2018prediction,pham2017hand,ehsani2020use,zhu2016inferring}. Due to the scarcity of real-world force data, most vision-based approaches rely on simulation \cite{ehsani2020use,zhu2016inferring,li2019estimating}. In robotics, a growing body of work addresses force-aware visual representations through real-world data collection \cite{chi2024multi,dave2024multimodal,yu2023mimictouch}, though these are confined to simple manipulation tasks such as pick-and-place. The closest work to ours is \cite{song2025opentouch}; our dataset is over $5\times$ larger and we explore tasks un-explored within \cite{song2025opentouch}.

\textbf{Applications of Force and Contact Understanding}
Reliably inferring physical hand-object interactions unlocks a range of important downstream tasks. Contact understanding has been shown to improve action recognition \cite{dessalene2021forecasting,shiota2024egocentric,liu2022joint} and to enable robots to learn manipulation trajectories by parsing contacts between hands and objects from human video \cite{singh2025hand,sivakumar2022robotic,pan2025spider}. Force understanding has seen comparatively fewer applications: \cite{adeniji2025feel} trains force-aware robot policies on a small dataset of simplistic tasks, and \cite{song2025opentouch} successfully demonstrates that forces improve grasp classification performance. We are the first to demonstrate zero-shot transfer of our learned representations over our force dataset for both contact understanding and action representation learning.
\section{Methods}
\label{sec:methods}

We introduced two families of learning tasks that leverage force as supervision: \textbf{contact understanding} and \textbf{action representation learning} across modalities. Contact understanding jointly addresses temporal contact segmentation (per-frame contact state classification) and pixel-level segmentation of objects in contact. For action understanding, we use force prediction as a pretraining objective to learn representations that are more physically grounded than appearance based features alone. Figure~\ref{fig:task_setup} illustrates our approach: we train modality-specific backbones on image and video inputs with force prediction as a self-supervised pretraining objective. 

The remainder of this section is organized as follows: Section~\ref{sec:dataset_collection} describes our data collection pipeline and the custom force-sensing hardware; Section~\ref{sec:contact_understanding} details contact understanding; Section ~\ref{sec:video_forces} describes action representation learning. The availability of synchronized force measurements as a new supervisory modality enables both the scalable derivation of contact annotations and the learning of physically grounded action representations — applications that would be difficult or impossible with vision alone.

% The remainder of this section is organized as follows: Section~\ref{sec:dataset_collection} describes our data collection pipeline and the custom force-sensing hardware; Section~\ref{sec:contact_understanding} details our force-supervised contact modeling approach; Section ~\ref{sec:video_forces} describes how we learn force-aware representations from video. The availability of synchronized force measurements as a new supervisory modality enables both the scalable derivation of contact annotations and the learning of physically grounded action representations — applications that would be difficult or impossible with vision alone.

\begin{figure}[b!]
    \centering
    \includegraphics[width=1.0\textwidth]{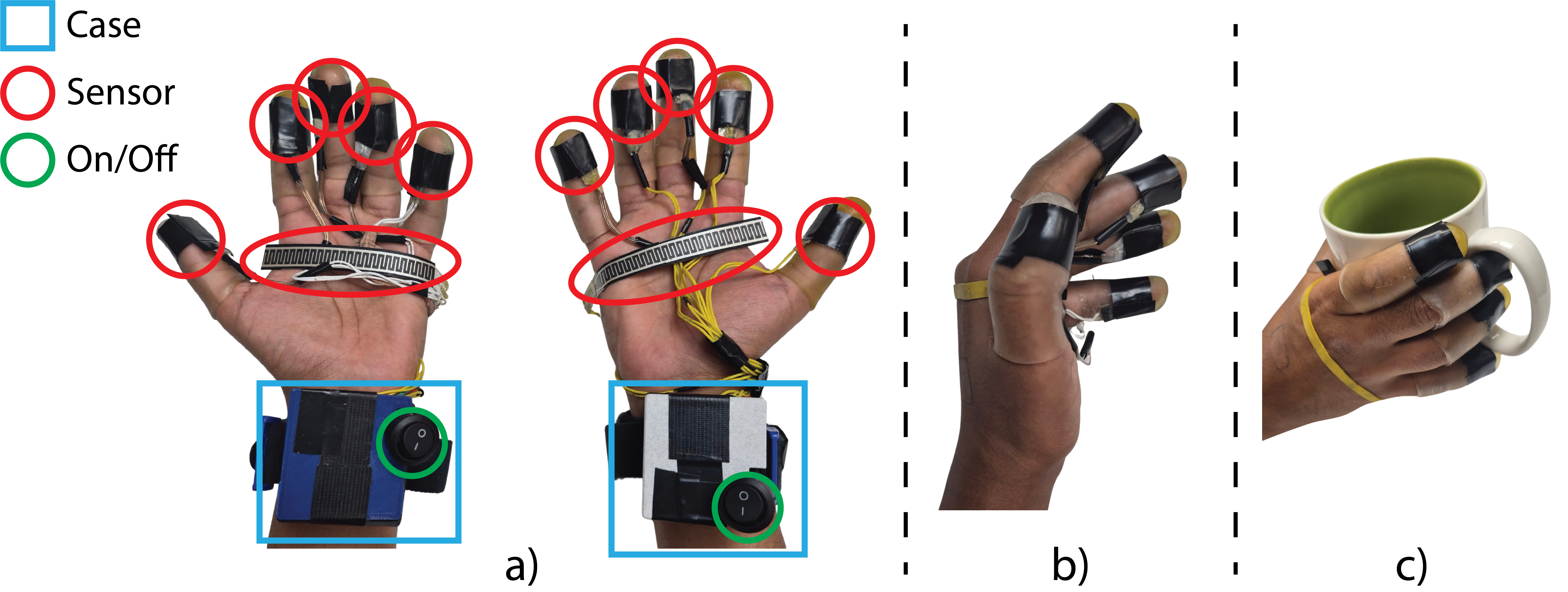}
    \caption{\textbf{Force-sensing glove hardware.} Leftmost: Palm-facing views of the left and right gloves, showing the six piezoresistive sensors ({\color{red} red}) mounted at each fingertip and across the palm, the Arduino microcontroller case ({\color{cyan} blue}) mounted at the wrist, and the on/off switch ({\color{ForestGreen} green}). Middle and right: Side and in-use views demonstrating the glove's low-profile form factor during natural object manipulation.}
    \label{fig:hardware}
\end{figure}

% Key pionts for Figure:
% -Multiple tasks: action recognition, action localization (fully supervised)
%                  force and contact prediction (self supervised)

% -Multiple modalities: images, audio, video
% -Selfsupervised traiing
% -Contact segmentation
% -Generic plug in architecture
% -Force as a pre-training task beneficial for both: 1) contact segmentation, 2) action understanding

% Figure 4) (methods) - Eadom
% Contact segmentation from video
% multimodal action understanding
% self-supervised learning across different modalities for downstream multimodal action understanding

% Cover both still image and video input
% Leave architecture generic
% Focus on modalities (Images, Video, Audio, Forces)

% Predicting forces is useful and produces:
%     Contact segmentation (Image -> Forces)
%     Multimodal action understanding (Video -> Forces, Video + Audio -> Forces)

% Show force predictino head after models, make it clear that it is additional training signal

% Show backbone + force prediction head trained:
%     force prediction head is not discarded for contact segmentation
%     force prediction head discarded for multimodal action understanding, use force-aware backbone

% Show data on left, and things to do with data on right, but not show data going into methods

% $\odv{f_r}{t}$

\begin{figure}[t!]
    \centering
    \includegraphics[width=0.66\textwidth]{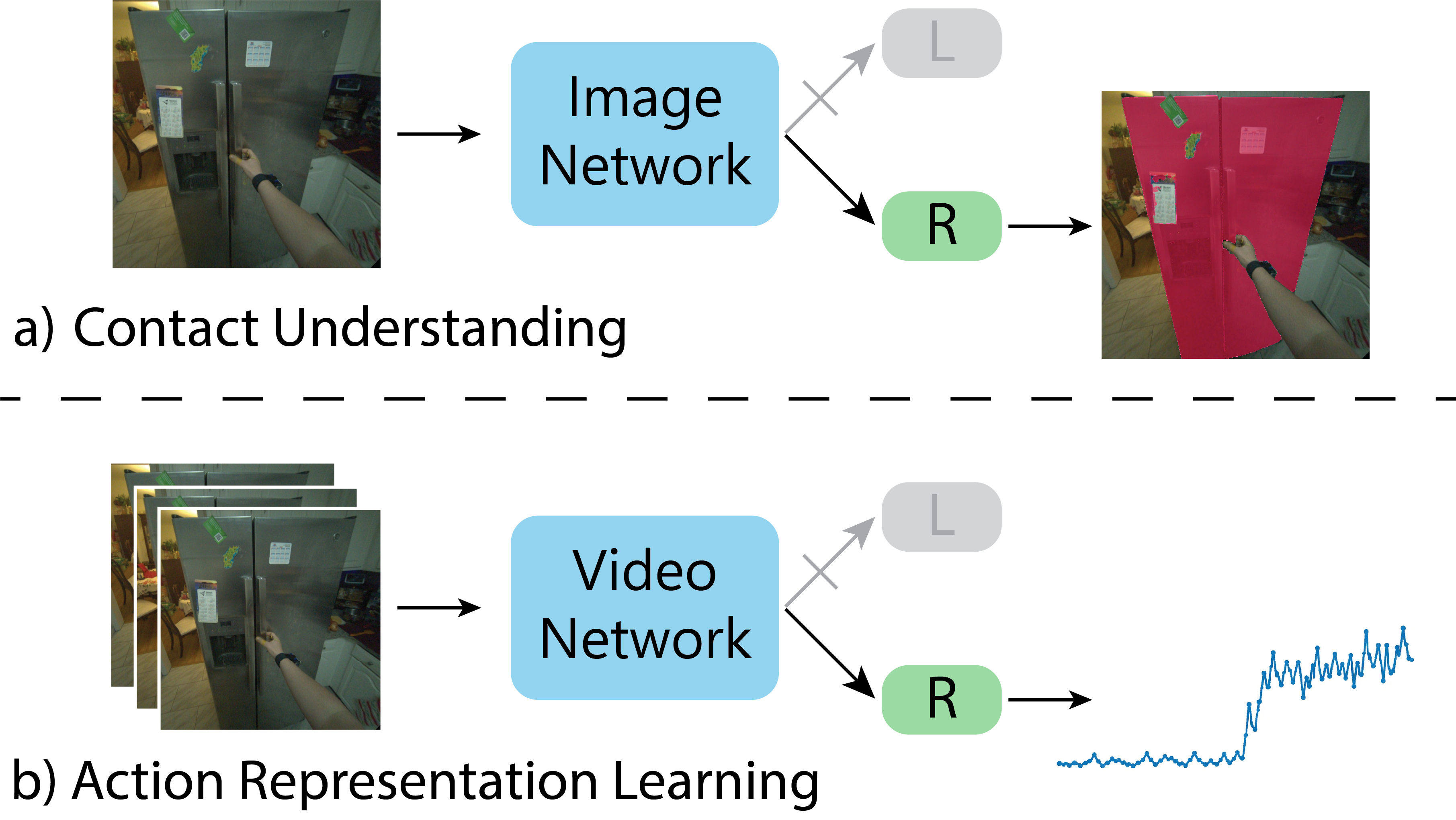}
    % \caption{We train modality-specific backbones on images, audio, and video, using synchronized force signals as supervision without manual labels. We demonstrate applications of FEEL over two popular downstream tasks: (i) for images we model contact understanding, where image features are used for contact detection and pixel-level segmentation of the object(s) in contact; and (ii) for audio and video we pre-train over forces directly, then transfer the force-aware pre-trained representations to fully supervised action recognition and detection benchmarks over popular video datasets.}
        \caption{\textbf{Learning contacts and action from forces.} \textbf{(a) Contact Understanding} Image network trained with force-derived contact labels performs contact detection and segmentation (red mask) for each hand (L/R).  \textbf{(b) Action Representation Learning} Video network predicts per-hand forces from clips for pretraining. The video backbone is transferred to action recognition after discarding force heads. No manual contact or action labels required during pre-training.}
        %add vertical space between abc labels and the images, make abc font larger, make L components and arrow going into L all grey, make opaqueness less to make it appear faded, make video vertically aligned with image and spectrogram, make image and segmented image crops identical, different arrows are of different weights and sizes (and other arrows), make force plot wider, higher, and thicker, make model boxes larger to allow text inside, same with L and R, vertical misalingment between abc, add intermediary component "LR" with combined color of purple and green and vertically align outputs
    \label{fig:task_setup}
\end{figure}
%change a) to a) contact understanding and b) action representation learning.

\subsection{Dataset Collection}
\label{sec:dataset_collection}
% {\color{blue} Start by structuring around hardware and hardware figure}
To obtain synchronized visual, acoustic, and force streams, we developed a lightweight sensing system combining a custom piezoresistive glove with Meta Aria glasses for egocentric capture.

\textbf{Hardware} Our sensing system combines two synchronized components: a custom piezoresistive force glove and Meta's Project Aria glasses (Gen 1) . This pairing enables simultaneous capture of manipulation forces, egocentric visual streams, and spatial audio.

\textbf{Force Glove} The force sensing component consists of six piezoresistive sensors (Flexiforce A301, Tekscan) mounted on a glove - one sensor per fingertip and a sensor strip mounted on the palm. See Figure \ref{fig:hardware} for the glove we construct and use for data collection. Each A301 sensor measures forces up to $45$ N (approximately $10$ lbs). The sensors connect to an Arduino Nano microcontroller mounted in a case enclosure that also houses a $3.3V$ Lithium Polymer (LiPo) battery and toggle switch for powering the device on and off to start and stop recordings. 

% Wiring is routed along the underside of the hand to minimize interference with natural grasping motions. 

% —a research platform with a comprehensive sensor suite while maintaining lightweight form factor {\color{blue} make sure not plagiarized}

\textbf{Project Aria Glasses} For egocentric visual stream and spatial audio capture, we use Meta's Project Aria device. The device captures synchronized multimodal streams: a forward-facing RGB camera ($1440\times1440$ pixels), two monochrome scene cameras ($640\times480$ pixels) positioned on the left and right temples for wide peripheral coverage, a 7-microphone array for spatial audio, and dual IMUs sampled at $800$ Hz and $1000$ Hz.

\textbf{Sensor Calibration and Synchronization} Synchronizing the force gloves and Aria glasses requires aligning two independent data streams operating on separate internal clocks. See the Supplementary Materials for the three-step calibration procedure employed at the start of each recording session.

\begin{table*}[t]
\centering
\caption{\textbf{Comparison of commonly used hand-object interaction datasets.} Modality codes: V=video, A=audio, D=depth, F=force, E=eye tracking. Contact annotation for Ego4D (FHO) is partial (only a single critical frame is annotated out of each sequence). FEEL is the only real-world (Real) egocentric dataset providing both contact annotations and force measurements — all other datasets with force lack contact labels or real-world capture, and all datasets with contact lack force.}
\label{tab:dataset_comparison}
\setlength{\tabcolsep}{5pt}
\renewcommand{\arraystretch}{1.15}
\begin{tabular}{l l l c c c c}
\hline
\textbf{Dataset} & \textbf{View} & \textbf{Scale} & \textbf{Real} & \textbf{Modalities} & \textbf{Contact} & \textbf{Forces} \\
\hline
EgoHands \cite{lin2020ego2hands} & Ego & 1.5 hrs & \checkmark & V & $\times$ & $\times$ \\
Arctic \cite{fan2023arctic} & Both & 4 hrs & \checkmark & V/D & \checkmark & $\times$ \\
VISOR \cite{darkhalil2022epic} & Ego & 36 hrs & \checkmark & V/A & \checkmark & $\times$ \\
HOI4D \cite{liu2022hoi4d} & Ego & 45 hrs & \checkmark & V/D & \checkmark & $\times$ \\
Ego4D (FHO) \cite{grauman2022ego4d} & Ego & 120 hrs & \checkmark & V/A & 1/2  & $\times$ \\
100DOH \cite{shan2020understanding} & Both & - & \checkmark & V & \checkmark & $\times$ \\
Zhang et al. \cite{zhang2025force} & Neither & 1 hour & $\times$ & V/F & $\times$ & \checkmark \\
Ehsani et al. \cite{ehsani2020use} & Exo & 0.5 hrs & $\times$ & V/F & \checkmark & \checkmark \\
OpenTouch. \cite{song2025opentouch} & Ego & 5 hrs & \checkmark & V/A/F/E & \checkmark & \checkmark \\
 \textbf{FEEL (Ours)} & \textbf{Ego} & \textbf{27 hrs} & \textbf{\checkmark} & \textbf{V/A/F/E} & \textbf{\checkmark} & \textbf{\checkmark} \\
\hline
\end{tabular}
\end{table*}
%list of modalities instead of single modality and check mark on ours
%scale, remove annotations
%action understanding vs robotics
%real world
%contact
%force
%  takeaway: FEEL is the only force-grounded contact dataset (all other force datasets don't havel contact, and all other contact datasets don't ground it in force)
%remove times and restructure table

%center green circles on figure
%separate table showing modalities and size of data, comparison between different datasets (scripted vs unscripted, set ontology vs open ontology)
%move legend away from hand
%center green on on/off (right hand)
%veritcally align vertical lines
%add a) b) c)

\textbf{Data Processing}
% \label{sec:data_processing}
\textbf{Egocentric Pose and Hand Tracking} We leverage Meta's Machine Perception Services (MPS) to extract foundational geometric and kinematic information from the Aria recordings. MPS processes the side camera streams and IMU data through a visual-inertial odometry pipeline to produce high-frequency (1 kHz) 6-DoF camera trajectories. Additionally, MPS performs 3D hand tracking on the side camera imagery, recovering hand keypoints for each frame in the recordings.

\textbf{Additional Processing} To arrive at nouns of objects of interaction, we adopt a weak labeling strategy: for each recording session, we partition the video into $2$-minute temporal windows and select $8$ uniformly sampled frames within each window. Over those frames we annotate the set of object nouns interacted with, without specifying temporal order or precise timing. We further compute dense geometric cues to support contacted object segmentation (see Section \ref{sec:contact_understanding}): scaled depth maps are predicted (in meters) using MLDepthPro \cite{bochkovskii2024depth}, and optical flow fields are densely estimated between consecutive frames using SEA-RAFT \cite{wang2024sea}.

\subsection{Contact Understanding}
\label{sec:contact_understanding}

Prior approaches to contact understanding \cite{shan2020understanding,darkhalil2022epic} depend on human annotators to label contact events from RGB frames—a process that is both labor-intensive and difficult to scale. Our approach instead exploits direct physical measurements - which, among other advantages, enable us to determine contact states and temporally localize contacted objects without manual annotation. From these force-derived contact states, we introduce a weakly supervised method for generating pseudolabel segmentations of contacted objects. This physically grounded supervision enables training paradigms that would not be possible with vision-only approaches.

\textbf{Contact Detection} Raw piezoresistive sensors exhibit substantial baseline drift and high-frequency noise that preclude simple thresholding for contact detection. We apply a multi-stage filtering pipeline to each sensor: outlier removal via Hampel filtering, Gaussian smoothing, time-varying baseline estimation via rolling percentile, baseline subtraction with negative clipping, and exclusion of regions with abrupt baseline changes or high RMS magnitudes. We then normalize each sensor by its $99.5$th percentile value and compute the geometric mean across all six sensors to produce a unified force signal robust to individual sensor failures. Frames where this signal exceeds an upper threshold indicate contact, frames below a lower threshold indicate non-contact, and frames between thresholds are excluded as ambiguous regions. For full details see the Supplementary Materials.

\textbf{Segmentation Pseudolabel Generation} We develop a weakly supervised approach to generate pseudo-labels for spatially segmenting objects in contact, eliminating the need for manual object segmentation annotations.

The inputs to our segmentation pipeline are: (i) frames designated as containing contact by our force-based detection system; (ii) a set of object mask proposals produced by a set of concept prompts fed to SAM3 (the set of concepts are the only weak labels provided, produced as per Section \ref{sec:dataset_collection}); (iii) pre-computed  optical flow fields using SEA-RAFT; (iv) camera intrinsics/extrinsics from MPS; and (v) pre-computed depth maps. The process by which these inputs are derived is described in Section \ref{sec:dataset_collection}.

Given these inputs, we generate contacted-object pseudo-labels by:  \textbf{(1)} extracting the masked optical flow for each object proposal; \textbf{(2)} computing the fundamental matrix between the current frame and the next frame using the Aria Machine Perception Services (MPS) provided intrinsics/extrinsics; \textbf{(3)} evaluating the Sampson epipolar error \cite{luong1993determining} within each mask using the masked flow and fundamental matrix (where higher error indicates stronger violations of the static-scene (epipolar) constraint due to object motion or manipulation); \textbf{(4)} selecting the proposal with the maximum mean epipolar error; and \textbf{(5)} accepting this proposal as the contacted object if and only if at least a set number of pixels belonging to the mask lie within a 3D distance threshold of the 3D hand centroid (ensuring that the selected high-motion region is spatially proximate to the manipulating hand). We have $500K$ images with binary contact pseudolabels, $180K$ of which contain binary segmentations of the object(s) in contact. For full details see Supplementary Materials.

\textbf{Training} To demonstrate the utility of FEEL, we train a contact understanding model for comparison against state of the art vision networks trained against large datasets of manually collected contact annotations. As illustrated in the first row of Figure \ref{fig:task_setup}, we train this network to process a single RGB image and produce two prediction heads per hand (left and right): (1) a binary contact classifier indicating whether the hand is in contact with an object, and (2) a segmentor of the object in contact with that hand. We supervise the devoted network head for each hand only when that hand is at least partially visible in the frame; network heads corresponding to out-of-view hands receive no supervision. For data augmentation, we apply random crops and random horizontal flips. When flipping, we mirror both the input image and the target outputs in order to maintain correspondence.

\subsection{Learning Forces from Video}
\label{sec:video_forces}

% Video networks ingest a compact stack of frames which sparsely samples the temporal dimension. In part due to this sparsity, important interactions in the input may be visually subtle or not fully observed. Modeling the true causal dynamics of action (force) allows better interpolation than simply modeling of incidentally associated features (visual appearance) {\color{blue} (phrasing and framing connection back to paragraph 1)}. 
%  and interpolate physical action tied to unobserved frames from video

Traditionally, video networks are pretrained on large-scale datasets of labeled images and videos (e.g., ImageNet \cite{deng2009imagenet} and Kinetics \cite{carreira2017quo}), which provide strong appearance- and motion-level priors but little direct supervision about the underlying physical interaction; in contrast, we use FEEL’s physically grounded force signals to pretrain video models to capture force dynamics (see Figure \ref{fig:task_setup}) and transfer these representations to downstream action understanding tasks. We formulate video-to-force prediction as a self-supervised pretraining task, then transfer the learned representations to downstream action recognition benchmarks. During the video-to-force pretraining, we only select clips for which at least one hand is visible for 50\% of the frames, ensuring that a training signal is reflected in the input images.

\begin{table*}[t]
\centering
\caption{Temporal contact detection results evaluated on the validation sets of FEEL, EPIC-VISOR, and HOI4D. We emphasize that while our model was only trained over $1$ dataset (FEEL), the 100DOH model and VISOR model were both trained over the EPIC-VISOR dataset~\cite{darkhalil2022epic}. \textbf{BC} represents "Binary contact detection" - a prediction is correct iff it captures the binary contact state between the hand and the object, irrespective of which predicted hand side is involved. \textbf{BSC} represents "Binary Signed Contact" - a prediction is correct iff it captures both the binary contact state of between the hand and the object, as well as the hand side involved.}
\label{tab:contact_detection}
\setlength{\tabcolsep}{6pt}
\begin{tabular}{l l l cc cc cc}
\hline
\multirow{2}{*}{\textbf{Method}} & \multirow{2}{*}{\textbf{Sup.}} & \multirow{2}{*}{\textbf{Training Set \#}} &
\multicolumn{2}{c}{\textbf{HOI4D}} &
\multicolumn{2}{c}{\textbf{VISOR}} &
\multicolumn{2}{c}{\textbf{FEEL}} \\
\cline{4-9}
& & & \textbf{BC} & \textbf{BSC} & \textbf{BC} & \textbf{BSC} & \textbf{BC} & \textbf{BSC} \\
\hline
Depth \cite{bochkovskii2024depth} & Self & 5 & 66.6 & 66.0 & 52.8 & 39.9 & 48.6 & 30.0 \\
HOIRef~\cite{bansal2024hoi} & Full & 2 (inc. VISOR) & 52.1 & 50.8 & 55.4 & 53.7 & 59.6 & 57.2 \\
VISOR~\cite{darkhalil2022epic} & Full & 1 (VISOR) & 55.3 & 55.0 & 66.3 & 65.2 & 45.2 & 40.2 \\
100DOH~\cite{shan2020understanding} & Full & 2 (inc. VISOR) & \textbf{90.0} & 80.5 & 81.3 & 71.9 & 49.0 & 40.0 \\
\hline
\textbf{Ours} & Self & 1 (inc. FEEL) & 82.2 & \textbf{81.3} & \textbf{87.3} & \textbf{79.7} & \textbf{95.6} & \textbf{88.9} \\
\hline
\end{tabular}
\end{table*}

\begin{table*}[t]
\centering
\caption{Spatial contact segmentation results evaluated on the validation sets of FEEL, EPIC-VISOR, and HOI4D. We emphasize that while our model was only trained over $1$ dataset (FEEL), the 100DOH model and VISOR model were both trained over the EPIC-VISOR dataset. As the 100DOH model produces bounding boxes (not segmentations), we employ in addition a SAM model (trained over Internet-scale data) to segment the 100DOH model predictions. Results reported as intersection-over-union (IOU) $\uparrow$.}
\label{tab:contact_segmentation}
\setlength{\tabcolsep}{6pt}
\begin{tabular}{l l l ccc}
\hline
\multirow{2}{*}{\textbf{Method}} & \multirow{2}{*}{\textbf{Sup.}} & \multirow{2}{*}{\textbf{Training Set \#}} &
\multicolumn{3}{c}{\textbf{Dataset}} \\
\cline{4-6}
& & & \textbf{HOI4D} & \textbf{VISOR} & \textbf{FEEL} \\
\hline
MOVES~\cite{higgins2023moves} & Weak & N/A & - & - & 0.309 \\
SAM3~\cite{carion2025sam} & Full & Comp. & 0.201 & 0.085 & 0.030 \\
VISOR~\cite{darkhalil2022epic} & Full & 1 (inc. VISOR) & 0.502 & 0.402 & 0.111 \\
100DOH~\cite{shan2020understanding}+SAM\cite{carion2025sam} & Full & Comp. (inc. VISOR) & \textbf{0.765} & \textbf{0.443} & 0.218 \\
\hline
\textbf{Ours} & Weak & 1 (FEEL) & 0.751 & 0.399 & \textbf{0.625} \\
\hline
\end{tabular}
\end{table*}

\section{Experiments}
\label{sec:experiments}

We evaluate FEEL across two families of tasks: (1) \textbf{contact understanding}, where we perform temporal contact detection and pixel-level object segmentation, and (2) \textbf{action representation learning}, where force prediction serves as a self-supervised pretraining objective for video backbones. See Figure \ref{fig:task_setup} for a depiction of the modeling of both tasks.

\subsection{Contact Understanding}

% {\color{blue} Give explanation / definition of contact understanding as a task first: define input and output (see figure 4)}

\textbf{Task: Contact understanding} jointly addresses temporal contact segmentation (per-frame contact state classification) and pixel-level segmentation of objects in contact. The model takes an image as input and produces a prediction as to the binary contact state of each hand as well as a pixel-level segmentation of the contacted object for each hand. See Figure \ref{fig:task_setup}.

\textbf{Model and implementation:} Traditional contact understanding models \cite{shan2020understanding} are complex to train due to the non-differentiable operations that happen along the way, along with the additional hyperparameters introduced. For simplicity of implementation, we train a Dense Prediction Transformer (DPT) \cite{ranftl2021vision} on top of frozen DINOv3\cite{simeoni2025dinov3} features. The image is fed as input to DINO, which produces feature maps subsequently fed into DPT, which produces two sets of outputs: one set of outputs as a binary logit based on the contact detected for each of the left and right hand; and a two-channel segmentation map devoted to the left and right hand, returned at the input resolution of the image. For further implementation details, see the Supplementary Materials.

\begin{figure}[t!]
    \centering
    \includegraphics[width=1\textwidth]{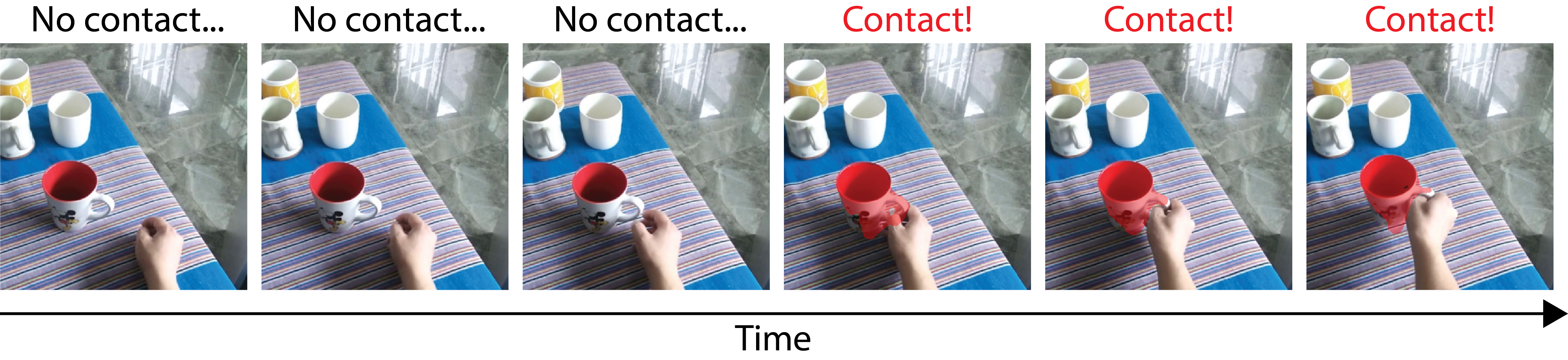}
    \caption{\textbf{Temporal contact detection.} A representative sequence showing our model's contact predictions over time as a hand reaches toward and grasps a mug. The model correctly predicts no contact during the approach phase and transitions to contact  precisely at the moment of grasp.}
    \label{fig:contact_temporal}
\end{figure}
%make uniform spacing horizontally
%zoom in into image between hand/obj
%add vertical padding between text and images (and imgs)
%add third dot to no contact
%reduce font size of time

\textbf{Baselines} The contact detections we learn over are derived in a fully self-supervised fashion; the contact segmentations we learn over are derived in a weakly-supervised fashion, as the pseudolabel generation algorithm is provided as input the object categories of all objects interacted with throughout the recording session. At no point is our training dependent on manually provided object segmentations. As such, we compare against other (1) self-supervised approaches, (2) weakly-supervised approaches, and (3) fully-supervised approaches.

\noindent\textbf{Contact detection baselines.} To our knowledge, there is no commonly used method for learning to detect contact in a self-supervised fashion. Therefore we introduce a simple zero-shot baseline combining recent advances in depth and segmentation networks. We compute segmentations of the body, and from the depth map of the scene compute a signed distance field of the scene with respect to each of the two hands. If the number of pixels where the distance $d(x, y)$ for all pixels $(x, y)$ exceeds a threshold, the contact prediction is positive, and vice-versa. We also compare against three fully-supervised baselines: HOIRef \cite{bansal2024hoi}, EPIC-VISOR \cite{darkhalil2022epic}, and 100DOH \cite{shan2020understanding}. See Table \ref{tab:contact_detection} for results.

\noindent\textbf{Contact segmentation baselines.} As for comparisons to other contact segmentation approaches, the closest self-supervised and weakly-supervised comparisons are \cite{higgins2023moves} and \cite{shan2021cohesiv}, but their methods are not publicly available. We instead compare against a reimplementation of the method for generating ground truth pseudolabels produced within \cite{higgins2023moves} - we compare our network's segmentation predictions directly against those pseudolabels. We also compare against SAM with its concept prompting, as well as two fully-supervised baselines: EPIC-VISOR and 100DOH. See Table \ref{tab:contact_segmentation} for results.

\begin{figure}[t!]
    \centering
    \includegraphics[width=1\textwidth]{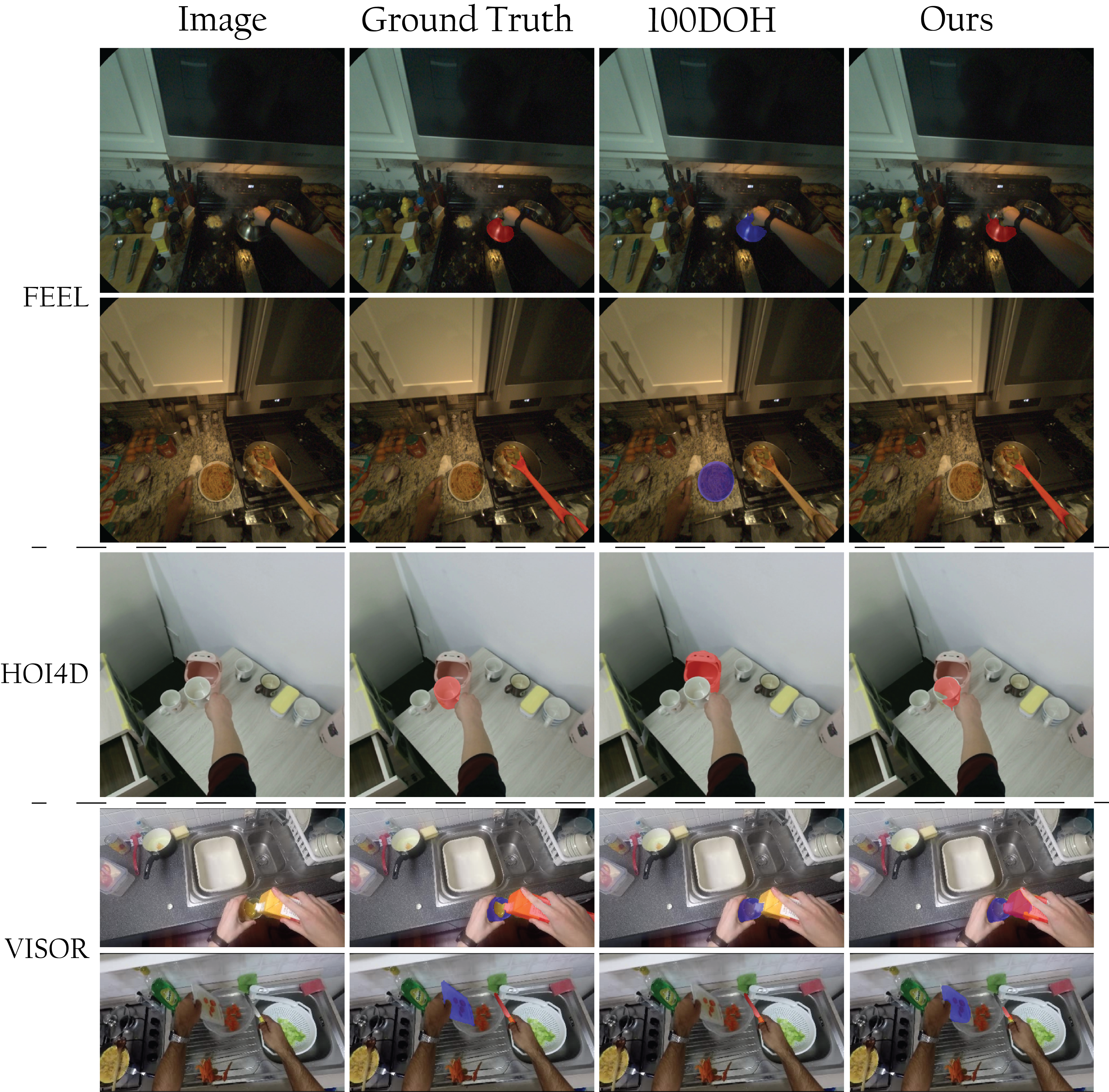}
    \caption{\textbf{Qualitative contact segmentation results.} Each row shows an input image, ground truth, 100DOH prediction, and our prediction across FEEL, HOI4D, and VISOR. Right and left hand contacts are shown in {\color{red} red} and {\color{blue} blue} respectively. Our model correctly identifies hand side (row 1), avoids confusing proximity for contact (rows 2-3), and handles simultaneous two-hand contacts (rows 4-5). Where our model produces partial segmentations (row 3), 100DOH (combined with SAM) confidently segments the wrong object entirely.}
    \label{fig:contact_segmentation}
\end{figure}

%closer crops verticallly (maybe)
%add borders to each image and add dividers between images
%add dataset label to the left side
%add divider between different dataset
%add space between text and images

\textbf{Evaluation Datasets}
We evaluate our method as well as other method over the contact pseudolabels belonging to FEEL, as well as zero-shot transfer over two popular egocentric video datasets that contain contact annotations - EPIC VISOR \cite{darkhalil2022epic} and HOI4D \cite{liu2022hoi4d}. 

%We attempted to perform evaluation over Ego4D \cite{grauman2022ego4d} but as of February 2026 the download link is not functioning\footnote{https://github.com/facebookresearch/Ego4d/issues/415}. When the link is opened we will perform our evaluation over that dataset.

\begin{table*}[t]
\centering
\caption{\textbf{Action recognition results across all benchmarks.} Top-1 accuracy for frozen and unfrozen settings across EPIC-Kitchens (verb/noun), SSV2, Ego-Exo4D, and Meccano. For each model, we compare standard pretraining (K710 for Hiera-B, Ego4D for EgoVideo-B) against the same weights further pretrained on FEEL's force data. Bold denotes the better result within each model pair. Models pre-trained over FEEL's data generally outperform the models pre-trained solely over Kinetics and Ego4D. We particularly observe this in the frozen evaluation setting compared to the unfrozen, end-to-end training.}
\label{tab:action_recognition}
\setlength{\tabcolsep}{5pt}
\begin{tabular}{l l l cc c c c}
\hline
& \multirow{2}{*}{\textbf{Model}} & \multirow{2}{*}{\textbf{Pretrain}} &
\multicolumn{2}{c}{\textbf{EPIC-Kitchens}} &
\textbf{SSV2} &
\textbf{Ego-Exo4D} &
\textbf{Meccano} \\
\cmidrule(lr){4-5} \cmidrule(lr){6-6} \cmidrule(lr){7-7} \cmidrule(lr){8-8}
& & & \textbf{Verb} & \textbf{Noun} & \textbf{Action} & \textbf{Action} & \textbf{Action} \\
\hline
\multirow{4}{*}{\rotatebox{90}{\textit{Frozen}}}
& Hiera-B    & K710  & 41.65          & 15.12          & 36.08          & 13.11          & 26.42 \\
& Hiera-B    & Ours  & \textbf{46.50} & \textbf{21.38} & \textbf{39.04} & \textbf{20.83} & \textbf{29.57} \\
\cline{2-8}
& EgoVideo-B & Ego4D & 59.24          & 29.42          & 49.61          & 26.00          & 36.59 \\
& EgoVideo-B & Ours  & \textbf{62.00} & \textbf{30.11} & \textbf{54.34} & \textbf{27.75} & \textbf{38.23} \\
\hline
\multirow{4}{*}{\rotatebox{90}{\textit{Unfrozen}}}
& Hiera-B    & K710  & \textbf{68.34} & 48.18          & 68.88          & 35.25          & 38.52 \\
& Hiera-B    & Ours  & 68.24          & \textbf{49.08} & \textbf{70.01} & \textbf{41.19} & \textbf{41.13} \\
\cline{2-8}
& EgoVideo-B & Ego4D & 70.99          & \textbf{55.67} & 74.55          & \textbf{40.92} & 48.99 \\
& EgoVideo-B & Ours  & \textbf{71.34} & 55.60          & \textbf{74.91} & 40.75          & \textbf{50.01} \\
\hline
\end{tabular}
\end{table*}

\subsection{Action Representation Learning from Video}

% {\color{blue} Give explanation / definition of contact understanding as a task first: define input and output (see figure 4)}
\textbf{Action representation learning} involves the exploring of pretraining objectives for video backbones before transferring of action representations for downstream tasks. When pre-training over FEEL, the model takes a video as input and produces force predictions. See Figure \ref{fig:task_setup}. We then transfer those learned representations to the task of action recognition. Here action recognition takes video as input, and produces a distribution over action categories as output. 

Due to computational constraints, rather than jointly pretraining on K710/Ego4D and FEEL, we instead initialize from K710 and Ego4D pretrained weights, fine-tune over FEEL's force data, and transfer the resulting force-aware representations to action recognition.

\textbf{Downstream Action Recognition} Forces are most impactful for the understanding of \textit{physical actions} performed in video. Therefore, after pretraining these video networks over FEEL, we focus our evaluation over downstream datasets centered on manipulation between hands and objects. The datasets we evaluate over are EPIC-Kitchens-100, SomethingSomething-V2, and Ego-Exo4D. We evaluate each model over the task of action recognition - in the case of EPIC-Kitchens, predicting \textit{verbs} and \textit{nouns}, whereas in Ego-Exo4D, SomethingSomething-V2 and Meccano the task is to predict \textit{actions}. For results, see Table \ref{tab:dataset_comparison}.

\section{Discussion}
\label{sec:discussion}
\textbf{Temporal Contact Detection} Qualitative results show in Figure 6. Note that our method accurately localizes the precise point in time at which contact is established, and is not confounded by mere hand / object proximity, correctly transitioning from no-contact to contact exactly at the onset of grasp formation. Table~\ref{tab:contact_detection} shows that our force-supervised model substantially outperforms the zero-shot depth baseline and surpasses fully supervised competitors across all three datasets, despite never receiving manually annotated contact labels. The most competitive baseline is 100DOH~\cite{shan2020understanding}, which benefits from training on two large, fully labeled datasets — the original 100DOH dataset and EPIC-VISOR — to achieve strong generalization to egocentric video. Our model surpasses the 100DOH model while relying solely on physical force measurements for supervision.

% For a qualitative example of our model's temporal precision, Figure~\ref{fig:contact_temporal} illustrates the model correctly transitioning from no-contact to contact exactly at the onset of grasp formation.

A further advantage of our approach is its handling of handedness. Our model experiences a smaller drop between the Binary Contact (BC) and Binary Signed Contact (BSC) metrics than 100DOH, reflecting more reliable left/right hand discrimination. The larger BC–BSC gap in 100DOH may stem from annotation noise: distinguishing hand side is subtle and error-prone for human annotators, potentially introducing label inconsistencies that degrade signed predictions. For ambiguities in annotations of handedness, see the Supplementary Materials.

\textbf{Contact Segmentation} Table~\ref{tab:contact_segmentation} demonstrates that our weakly supervised model achieves competitive spatial segmentation performance across all three datasets. On VISOR, our model slightly trails 100DOH, which is expected given that 1) the 100DOH model relies on SAM3 \cite{carion2025sam}, a powerful segmentation model trained over Internet-scale data, and 2) the 100DOH model is fully trained on the EPIC-VISOR training split while our model is deployed zero-shot. Despite this disadvantage, the gap is narrow.

% Our model correctly identifies hand side, avoids confusing proximity for contact, and handles simultaneous two-hand contacts. A qualitative difference between the two methods is visible in Figure~\ref{fig:contact_segmentation}: 100DOH, which first predicts a bounding box before invoking SAM~\cite{carion2025sam} for segmentation, tends to produce complete but occasionally over-confident masks of background objects. Our model, which predicts segmentations end-to-end without requiring a secondary segmentation stage, more consistently identifies the correct contacted object while sometimes producing partial coverage of the object body. 

Our model correctly identifies hand side, avoids confusing proximity for contact, and handles simultaneous two-hand contacts. While 100DOH — which relies on bounding box predictions followed by SAM~\cite{carion2025sam} — produces complete but occasionally over-confident masks of the wrong object, our end-to-end model more consistently identifies the correct contacted object, sometimes at the cost of partial coverage. See Figure ~\ref{fig:contact_segmentation} for examples.
% This trade-off reflects the difference in supervision: bounding box–derived SAM prompts favor completeness, whereas our force-derived pseudolabels favor precision at the contact site.

% Left and right hand contacts are shown in {\color{red} red} and {\color{blue} blue} respectively. 

\textbf{Action Representation Learning} Table \ref{tab:dataset_comparison} reveals two consistent trends across all four downstream benchmarks across Hiera \cite{ryali2023hiera} and EgoVideo \cite{pei2024egovideo} models.

\textit{Force pretraining improves frozen representations more than fine-tuned ones.} Performance gains are consistently larger in the frozen-eval setting than under end-to-end fine-tuning. This indicates that force-aware features are more immediately transferable than conventional appearance-based pretraining weights, and suggests that force supervision instills a qualitatively different representational prior.

% rather than simply providing additional training signal that fine-tuning can compensate for.
\textit{Gains are largest on smaller downstream datasets.} The four evaluation datasets span a wide range of sizes (Meccano smallest, Ego-Exo4D second smallest, EPIC Kitchens second largest, and SomethingSomething-V2 largest), and the relative performance benefit of force pretraining is most pronounced at the smaller end of this spectrum. This is consistent with the hypothesis that physically grounded representations reduce the amount of labeled downstream data needed to learn effective action classifiers — a practically valuable property given the cost of large-scale action annotation.

% (fine tuning is more challenging with smaller datasets, making the transferability benefit of force representations more pronounced)

% Taken together, these results suggest that force, as a low-dimensional causal signal, provides a complementary and non-redundant source of supervision to appearance- and motion-based pretraining, and that its benefits are most pronounced precisely where labeled data is scarce.

\textbf{Limitations.} On the hardware side, rubber gel pad mounts were uncomfortable for extended wear, individual sensors were sometimes unreliable due to mechanical drift and bending, and the sensors obstruct the fingertips during tasks requiring fine dexterity - limitations that could be addressed by moving to above-site sensing such as wrist-mounted EMG devices. On the algorithmic side, our pseudolabel generation relies on object motion, meaning we cannot segment stationary contacted objects such as tables or heavy appliances, which will require cues beyond motion to resolve.

\textbf{Summary} Force is the underlying cause that drives physical interaction, it is a critical primitive for physical action understanding. We release FEEL the largest egocentric force–video dataset, capturing the physical causes - not just the visual effects - of hand-object manipulation. We empirically demonstrate the utility of feel over both Contact Understanding, and Action Representation Learning without any need need for manual action label or object segmentation annotations. We hope FEEL serves as a foundation for future work spanning finer-grained hand understanding, force-aware visuomotor control, and manipulation policy learning from human video; and that as wearable sensing hardware continues to improve in form factor and comfort, the community will scale datasets like FEEL to further close the gap between the richness of physical interaction and what can be observed from video alone — with applications reaching from robotics and human-robot interaction to virtual and augmented reality.

\bibliographystyle{splncs04}
\bibliography{main}
\end{document}